%% file: main.tex
\documentclass[10pt,twocolumn,letterpaper]{article}

\usepackage{cvpr}              


\definecolor{cvprblue}{rgb}{0.21,0.49,0.74}
\usepackage[pagebackref,breaklinks,colorlinks,allcolors=cvprblue]{hyperref}
\usepackage{multirow}
\usepackage{pifont}
\def\paperID{*****} 
\def\confName{CVPR}
\def\confYear{2025}

\title{Cross-Spectral Body Recognition with Side Information Embedding: Benchmarks on LLCM and Analyzing Range-Induced Occlusions on IJB-MDF}

\author{Anirudh Nanduri\thanks{Equal contribution.}\\
University of Maryland\\
College Park, MD, USA\\
{\tt\small snanduri@umd.edu}
\and
Siyuan Huang\footnotemark[1]\\
Johns Hopkins University\\
Baltimore, MD, USA\\
{\tt\small shuan124@jhu.edu}
\and
Rama Chellappa\\
Johns Hopkins University\\
Baltimore, MD, USA\\
{\tt\small chella4@jhu.edu}
}

\begin{document}
\maketitle
\input{sec/0_abstract}    
\input{sec/1_intro}
\input{sec/2_formatting}
{
    \small
    \bibliographystyle{ieeenat_fullname}
    \bibliography{main}
}


\end{document}

%% file: sec/0_abstract.tex
\begin{abstract}
Vision Transformers (ViTs) have demonstrated impressive performance across a wide range of biometric tasks, including face and body recognition. In this work, we adapt a ViT model pretrained on visible (VIS) imagery to the challenging problem of cross-spectral body recognition, which involves matching images captured in the visible and infrared (IR) domains. Recent ViT architectures have explored incorporating additional embeddings beyond traditional positional embeddings. Building on this idea, we integrate Side Information Embedding (SIE) and examine the impact of encoding domain and camera information to enhance cross-spectral matching. Surprisingly, our results show that encoding only camera information—without explicitly incorporating domain information—achieves state-of-the-art performance on the LLCM dataset. While occlusion handling has been extensively studied in visible-spectrum person re-identification (Re-ID), occlusions in visible-infrared (VI) Re-ID remain largely underexplored — primarily because existing VI-ReID datasets, such as LLCM, SYSU-MM01, and RegDB, predominantly feature full-body, unoccluded images. To address this gap, we analyze the impact of range-induced occlusions using the IARPA Janus Benchmark Multi-Domain Face (IJB-MDF) dataset, which provides a diverse set of visible and infrared images captured at various distances, enabling cross-range, cross-spectral evaluations.
\end{abstract}

%% file: sec/1_intro.tex
\section{Introduction}
\label{sec:intro}

In this paper, we examine the problem of cross-spectral body-recognition across diverse ranges using Vision Transformer (ViT) architectures. When body recognition is formulated as a person re-identification (Re-ID) problem, the objective is to match identities across images or videos captured from multiple non-overlapping cameras. Several inherent challenges in this task include occlusion, pose variations, and viewpoint changes. These challenges become even more pronounced when coupled with domain shifts that arise in cross-spectral matching scenarios.

Vision Transformers have shown strong performance across various general computer vision tasks, such as image classification, object detection, and image segmentation, and have also been successfully applied to person Re-ID, including visible-infrared person Re-ID (VI-ReID). Transformers have been shown to effectively capture global shape and structural information, and demonstrate strong robustness to severe occlusions, perturbations, and domain shifts \cite{naseer2021intriguing}. Furthermore, the self-attention mechanism in ViTs enables local cross-modal interactions at the patch level, which is particularly beneficial for VI-ReID.

Although occlusion has been widely studied in the context of person identification, it has not been specifically explored in VI-ReID, primarily due to the absence of evaluation protocols and datasets that include significant occlusions. Existing VI-ReID benchmark datasets, such as RegDB, SYSU-MM01, and LLCM, lack substantial occlusions, limiting the study of occlusion-aware models in cross-spectral settings.

\begin{figure*}[h!]
	\centering
	\includegraphics[width=0.9\linewidth]{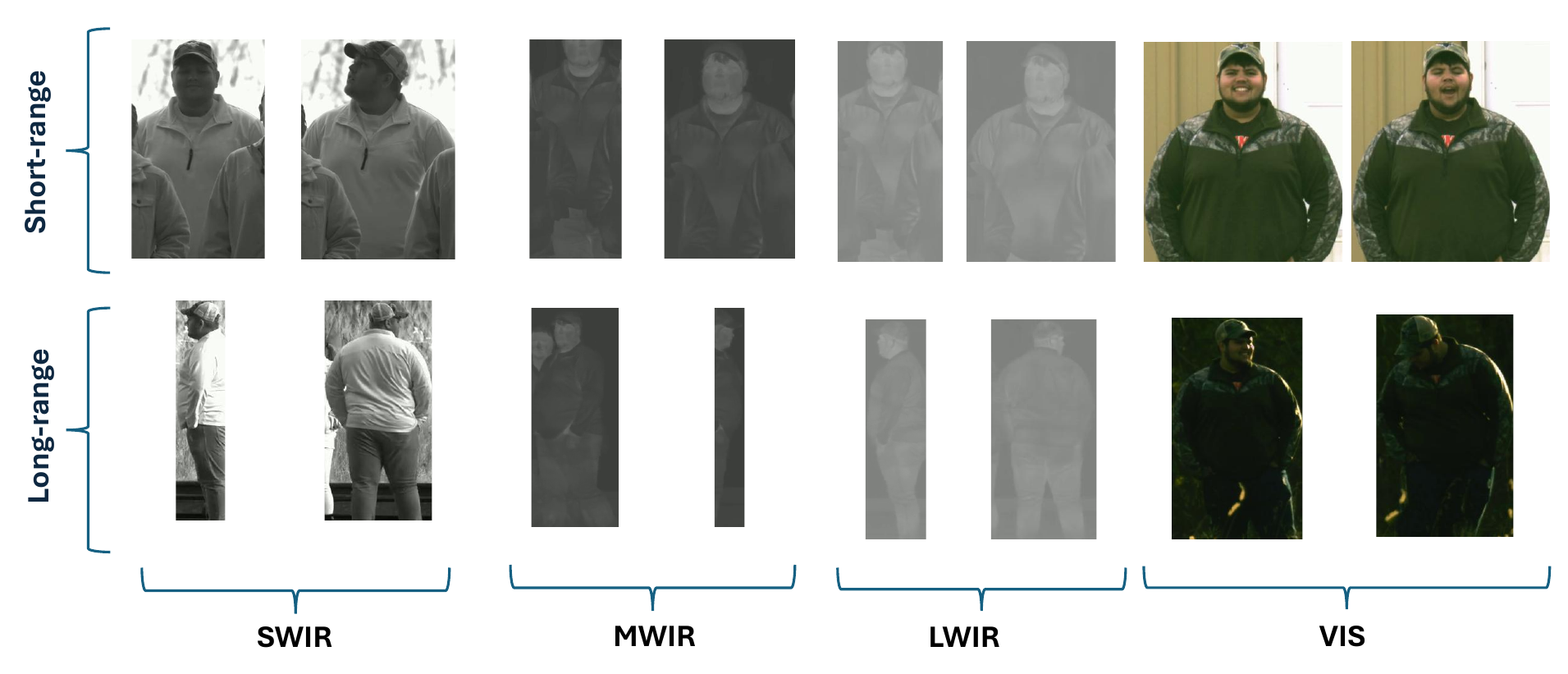}
	\caption{Example images from IJB-MDF dataset from VIS, SWIR, MWIR and LWIR domains captured at short- and long-ranges. The VIS images are captured at 300m and 500m. The IR images are captured at 15m and 30m.}
	\label{fig:c5_example_persons}
\end{figure*}

The IARPA Janus Benchmark Multi-Domain Face (IJB-MDF) dataset addresses some of these limitations by capturing visible imagery at long ranges (300m, 400m, and 500m) and infrared imagery at closer ranges (15m and 30m). As illustrated in Fig.~\ref{fig:c5_example_persons}, this diversity in capture ranges introduces significant scale variations, which inherently result in occlusions in the form of scale misalignment.

Through experiments conducted with a state-of-the-art ViT architecture, we demonstrate that the model is highly sensitive to scale and range variations, underscoring the need for models that can robustly handle cross-spectral and cross-range body recognition.

%% file: sec/2_formatting.tex
\section{Related Work}
\label{sec:related_work}


    
\subsection{VI Re-ID Methods:}

Most VI-ReID methods focus on learning modality-invariant features by disentangling modality-specific and modality-shared information. Zhang et al. \cite{zhang2023diverse} proposed DEEN, a Diverse Embedding Expansion Network that generates diverse embeddings to effectively reduce the domain gap between visible and infrared images. Wu et al. \cite{wu2024wrim} introduced WRIM-Net, which extracts modality-invariant features through a Multi-dimension Interactive Informative Mining (MIIM) module combined with a contrastive loss (CM-KIC). Ren et al. \cite{ren2024implicit} presented IDKL, a framework that leverages implicit discriminative knowledge within modality-specific features to enhance the discriminative capacity of shared features. More recently, Alehdaghi et al. \cite{alehdaghi2025cross} proposed MixER (Mixed Modality Erased and Related), which disentangles modality-specific and modality-shared identity information through a combination of orthogonal loss, modality-erased identity loss, modality-aware loss, and mixed cross-modal triplet loss functions.

\textbf{Transformer-based Approaches to VI Re-ID: }

Jiang et al. \cite{jiang2022cross} proposed the Cross-Modality Transformer (CMT), which incorporates a modality-level alignment module to compensate for missing modality-specific information and an instance-level alignment module that adaptively adjusts sample features via query-adaptive feature modulation. Chen et al. \cite{chen2022structure} introduced the Structure-aware Position Transformer (SPOT), which learns semantic-aware, shareable modality features by leveraging both structural and positional information. Feng et al. \cite{feng2022visible} presented the Cross-Modality Interaction Transformer (CMIT), which enhances patch-level feature representations through cross-modal patch token interactions and integrates local features from CNNs with global features from Transformers to capture salient cross-modal representations. Liang et al. \cite{liang2023cross} proposed the Cross-Modality Transformer (CMTR), which captures modality-specific information via modality embeddings fused with token embeddings, and introduced a modality-aware enhancement loss to improve the discriminability of learned embeddings.

\subsection{Occlusion in Person Re-ID:}

Occlusion in person Re-ID can arise from two primary sources: (1) upstream data processing following partial detections, and (2) physical obstructions during image capture. The first type of occlusion occurs when the detected bounding box contains only a part of the body, which, when scaled to fit the model's input size, leads to position and scale misalignment. The second type of occlusion results from external obstructions in the environment, where parts of the subject are blocked, leading to noisy or missing information in the image. 

Occlusions caused by spatial misalignment are typically handled using image transformation techniques and multi-scale feature extraction. In contrast, occlusions due to noisy or missing information are mitigated through feature alignment, attention mechanisms, and contextual recovery methods.

\textbf{Image transformation-based methods} aim to align partial images with holistic images. Some approaches transform partial images to resemble holistic ones \cite{zhong2020robust, he2021partial, lin2021partial, he2021seriouslyMisalign}, while others adjust regions from holistic images to match partial views \cite{luo2020partial, huo2021partial}.

\textbf{Multi-scale feature-based methods} capture both local and global information by operating at different scales. Some focus on local regions with varying receptive fields \cite{he2018partial, he2019pyramid, wang2021occluded}, while others utilize multi-scale pyramid features to integrate information across scales \cite{zheng2019pyramidal, yan2020videoreid}.

\textbf{Data augmentation-based attention methods} simulate occlusions through data augmentation and train the network in a self-supervised manner to enhance robustness to occlusions \cite{zhuo2018occluded, sun2019partial, zhong2020random, zhong2020robust, zhao2021incremental, yan2021occluded, chen2021occlude, jia2022learning, wang2022fed, ji2024exploring}. The problem addressed in this paper, where we perform transfer learning of a ViT model pretrained on holistic images, is most closely aligned with attention mechanism-based approaches, as both aim to adapt holistic models for handling partial occlusions effectively.

\section{Method}\label{sec:c5_method}

Given an input image $x \in \mathbb{R}^{H \times W \times C}$, where $H$, $W$, $C$ denote its height, width, and number of channels, respectively, we split it into $N$ patches $\{x^i_p|i=1,2,\cdots, N\}$ of size $P \times P$. These patches are then linearly projected to create patch embeddings $\mathcal{F}({x}^i_p)$ of dimension $D$. Extra learnable embedding token denoted as $x_\text{cls}$, and local embedding tokens that represent various local parts denoted as $x_\text{local}^j$ are prepended to the input sequences. Learnable position embeddings represented by $\mathcal{P}$ incorporate spatial information. Then, the input sequences fed into transformer layers can be expressed as:
\begin{multline}\label{eq:z01}
\mathcal{Z}_0 = [ {x}_\text{cls}; {x}_\text{local}^1; {x}_\text{local}^2; \cdots; {x}_\text{local}^K; \\
\, \mathcal{F}({x}^1_p) ; \, \mathcal{F}({x}^2_p); \cdots; \, \mathcal{F}({x}^{N}_p) ] + \mathcal{P},
\end{multline}

where $\mathcal{Z}_0$ represents input sequence embeddings and $\mathcal{P} \in \mathbb{R}^{(N + K + 1) \times D}$ is position embeddings. $\mathcal{F}$ is a linear projection mapping the patches to $D$ dimensions.

Inspired from side information embeddings (SIE) introduced by TransReID model \cite{he2021transreid}, we incorporate SIE embeddings to encode the domain and range information. Assuming we have a total of $N_D$ domains to encode, the side information embeddings can be represented by $\mathcal{S} \in \mathbb{R}^{N_D \times D}$. Then the domain specific embedding corresponding to the domain $i_{dom} \in \{0, 1, ..., N_D - 1\}$ is given by $\mathcal{S}[i_{dom}]$.

The final input to the transformer encoder is then represented as:
\vspace{-0.5em}
\begin{equation} \label{eq:sie}
	\mathcal{Z}_0^{'} = \mathcal{Z}_0 + \lambda \mathcal{S}[i_{dom}],
\end{equation}
where $\mathcal{Z}_0$ is defined as in Eq.~\ref{eq:z01}, and $\lambda$ is a hyperparameter to control the impact of SIE in the final input.


The output ${x}_\text{cls}$ token is used as the global feature representation, while the local semantic features are the output local tokens ${x}_\text{local}^i$ obtained from the ViT encoder. 

The final feature representation is obtained by concatenating the global feature with the mean of the local features:

\begin{equation} \label{eq:final_feature}
	\mathbf{f} = [\mathbf{f}_{\text{cls}}; \text{mean}(\mathbf{f}_{\text{local}}^1, \mathbf{f}_{\text{local}}^2, \cdots, \mathbf{f}_{\text{local}}^K)]
\end{equation}

We process images from all the domains similarly, except for the SIE $\mathcal{S}[i_{dom}]$. To encourage cross-spectral robustness, we form the mini-batches such that they have equal representation from all the domains under consideration. For each batch, we sample $P$ identities and then sample $K/N_D$ images of each identity, from each of the $N_D$ domains. Thus we end up with a batch-size of $PK$. This also facilitates an efficient implementation of triplet loss as discussed below.

We enforce identity-consistency using the cross-entropy loss. Given a batch of image features $\mathbf{f}_{\text{final}}$ generated from the model, the cross-entropy loss $\mathcal{L}_{\text{CE}}$ is given by,
\begin{equation}
	\mathcal{L}_{\text{CE}} = -\frac{1}{PK}\sum_{i=1}^{PK} \log \frac{\exp(\mathbf{f}_i^{y_i})}{\sum_{j=1}^{PK} \exp(\mathbf{f}_i^{y_j}))},
\end{equation}
where $PK$ is the batch size, and $y_i$ represents the identity label of $\mathbf{f}_i$.

The triplet loss \cite{hermans2017defense} formulation that includes selecting the hardest samples within the mini-batch is given by,

\begin{align}\label{eq:loss_bh}
        \mathcal{L}_{\text{tri}} = \sum\limits_{i=1}^{P} \sum\limits_{a=1}^{K}
            \Big[m & + \hspace*{-5pt} \max\limits_{p=1 \dots K} \hspace*{-5pt} D\left(\mathbf{f}^i_a, \mathbf{f}^i_p\right) \\
                   & - \hspace*{-5pt} \min\limits_{\substack{j=1 \dots P \\ n=1 \dots K \\ j \neq i}} \hspace*{-5pt} D\left(\mathbf{f}^i_a, \mathbf{f}^j_n\right) \Big]_+,\nonumber
\end{align}
where $m$ is the margin parameter, $D(\cdot,\cdot)$ measures the feature distance, and $[\cdot]_+$ denotes the hinge function.

The final training loss $\mathcal{L}$ is a weighted sum of the two losses:
\begin{equation}
	\mathcal{L} = \mathcal{L}_{\text{id}} + \lambda_t\mathcal{L}_{\text{tri}},
\end{equation}
where $\lambda_t$ is hyperparameter controlling the contribution of each loss term. 

\section{Experiments}\label{sec:c5_exp}
In this section, we present experimental results on the LLCM \cite{zhang2023diverse} and the IARPA Janus Benchmark Multi-Domain Face (IJB-MDF) \cite{kalka2019iarpa} dataset. 

\subsection{Datasets}

\textbf{Low-Light Cross-Modal (LLCM)} \cite{zhang2023diverse} is a benchmark dataset for VI-ReID, containing 46,767 images of 1,064 identities, captured using a 9-camera network in low-light environments. The dataset includes both VIS and IR images, covering diverse climate conditions and clothing variations. The training-to-test split is approximately 2:1, resulting in 30,921 bounding boxes (VIS: 16,946 and IR: 13,975) of 713 identities for training, and 21,075 bounding boxes (VIS: 13,909 and IR: 7,166) of 351 identities for testing. The LLCM dataset supports evaluation in both VIS-to-IR and IR-to-VIS matching modes.

\textbf{IJB-MDF} \cite{kalka2019iarpa} is a multi-domain dataset consisting of images and videos of 251 subjects, captured using a variety of cameras spanning visible, short-wave infrared (SWIR), medium-wave infrared (MWIR), long-wave infrared (LWIR), and long-range surveillance domains. The dataset includes 1,757 visible enrollment images, 40,597 SWIR enrollment images, and over 800 videos totaling 161 hours. It also provides ground-truth face bounding boxes, which we leverage in combination with YOLOv10-X \cite{THU-MIGyolov10} body detection results to generate body bounding-box ground-truth annotations. To assign identity labels to detected body bounding boxes, we match each detection to the face bounding box with which it has the maximum Intersection over Union (IoU). To minimize labeling errors, we discard any detection that has an IoU greater than 0.75 with more than one face bounding box.

We use subjects from gallery-1 for training and subjects from gallery-2 for testing. Videos from the following domains are considered: VIS-500m, SWIR-30m, MWIR-30m, LWIR-30m, VIS-300m, SWIR-15m, MWIR-15m, and LWIR-15m. As illustrated in Fig.~\ref{fig:c5_example_persons}, there is a significant variation in scale between long-range and short-range domains.

We construct the following three training datasets for our experiments:

\begin{itemize} \item \textbf{Short-range dataset}: VIS-300m, SWIR-15m, MWIR-15m, and LWIR-15m. \item \textbf{Long-range dataset}: VIS-500m, SWIR-30m, MWIR-30m, and LWIR-30m. \item \textbf{Mixed-range dataset}: VIS-300m, VIS-500m, SWIR-15m, SWIR-30m, MWIR-15m, MWIR-30m, LWIR-15m, and LWIR-30m. \end{itemize}

We propose the following test protocols for evaluating cross-spectral, cross-range body recognition:
\begin{itemize} \item \textbf{Short-range protocol}: Gallery set from \textit{VIS-300m}, and probe set from \textit{IR-15m}, \item \textbf{Long-range protocol}: Gallery set from \textit{VIS-500m}, and probe set from \textit{IR-30m}, \item \textbf{Mixed-range protocol 1}: Gallery set from \textit{VIS-300m}, and probe set from \textit{IR-30m}, \item \textbf{Mixed-range protocol 2}: Gallery set from \textit{VIS-500m}, and probe set from \textit{IR-15m}. \end{itemize}

In each test protocol, we randomly sample 10 images per identity for the gallery set and 100 images per identity for the probe set.

\subsection{Implementation Details}\label{sec:c5_implement}
We use a ViT-B model with a patch size of $16 \times 16$ that is first pretrained on the LUPerson dataset \cite{fu2020unsupervised} in a self-supervised manner as described in \cite{huang2023self}, without the side information embeddings. This allows the model to learn local semantic information about different body parts. The model is specifically trained with local part tokens that correspond to the face, torso and the lower body. 

We then finetune this model using a combination of cross-entropy and triplet losses, along with side information embeddings that encode the domain and/or camera information.

\subsubsection{LLCM}

The training batches are sampled such that each batch contains 16 subjects, with two images per subject from both the VIS and IR domains, resulting in a total batch size of 64. We employ a cosine learning rate scheduler with an initial learning rate of 0.0004 and a warm-up phase of 20 epochs. All images are resized to $384 \times 128$, and the following data augmentations are applied: random horizontal flip, padding, random crop, and random erasing \cite{zhong2020random}. The SIE parameter $\lambda$ in Equation~\ref{eq:sie} is set to $3$ and the triplet loss weight $\lambda_t$ is set to $1$. We use the SGD optimizer with momentum of $0.9$ and weight decay of $0.0001$. The model is trained for 120 epochs.
We perform experiments with three configurations of Side Information Embedding (SIE):
\begin{itemize} \item \textbf{SIE-2}: Only domain information is encoded, where \textit{VIS} $\rightarrow$ 0 and \textit{IR} $\rightarrow$ 1. \item \textbf{SIE-9}: Only camera information is encoded. Since the LLCM dataset contains a total of 9 cameras, each camera is assigned an index from 0 to 8. \item \textbf{SIE-18}: Both domain and camera information are encoded. The 9 \textit{VIS} cameras are mapped to indices 0--8, and the 9 \textit{IR} cameras are mapped to indices 9--17. \end{itemize}

\subsubsection{IJB-MDF}

Most of the training settings for IJB-MDF follow those described for LLCM, with the primary difference being in mini-batch formation. For both the short-range and long-range training sets, each batch contains 16 subjects, with one image per subject sampled from each of the VIS, SWIR, MWIR, and LWIR domains. For the mixed-range training set, each batch also consists of 16 subjects, but with two images per subject from each domain — one image from a short-range domain and one from a long-range domain — to ensure coverage of both range conditions within a batch. We perform experiments with two configurations of SIE:
\begin{itemize} 
\item \textbf{SIE-4}: Only domain information is encoded, where \textit{VIS} $\rightarrow$ 0, \textit{SWIR} $\rightarrow$ 1, \textit{MWIR} $\rightarrow$ 2 and \textit{LWIR} $\rightarrow$ 3. 
\item \textbf{SIE-8}: Both the range and domain information are encoded. The four short-range domains are mapped to indices 0--3, while the four long-range domains are mapped to indices 4--7.
\end{itemize}

\subsection{Results}

\subsubsection{LLCM}
\Cref{tab:reid_comparison} presents a comparison of our model's performance against recent state-of-the-art methods, including DEEN \cite{zhang2023diverse}, WRIM-Net \cite{wu2024wrim}, IRM (Instruct-ReID) \cite{he2024instruct}, IDKL \cite{ren2024implicit}, and MixER \cite{alehdaghi2025cross}. For IDKL, we report the results from \cite{alehdaghi2025cross}, which exclude re-ranking, to ensure a fair comparison with the other methods.

The results demonstrate that, even without explicitly encoding domain information, our ViT model trained with SIE-9 (which encodes only camera information) outperforms all other approaches. Furthermore, we observe that the joint encoding of both domain and camera information, as in SIE-18, results in a significant drop in performance. \textit{This highlights that, on the LLCM dataset, encoding camera information is more crucial than domain information for robust cross-modal matching.}

\begin{table*}[t!]
\renewcommand{\arraystretch}{1.5}
    \centering
    \begin{tabular}{l|cc|cc}
        \hline
        \multirow{2}{*}{\textbf{Model}} & \multicolumn{2}{c|}{\textbf{VIS to IR}} & \multicolumn{2}{c}{\textbf{IR to VIS}} \\
        
        & Rank-1 & mAP & Rank-1 & mAP \\
        \hline
        DEEN \cite{zhang2023diverse} & 62.5 & 65.8 & 54.9 & 62.9 \\
        WRIM-Net \cite{wu2024wrim} & 67.0 & 69.2 & 58.4 & 64.8 \\
        IRM (STL) \cite{he2024instruct} & 64.9 & 64.5 & \textbf{66.2} & 66.6 \\
        IDKL* \cite{ren2024implicit}  & 70.4 & 55.0 & 62.5 & 49.3 \\
        MixER \cite{alehdaghi2025cross} & 70.8 & 56.6 & 65.8 & 51.1 \\
        \hline
        Our Model & 70.5 & 73.2 & 63.9 & 70.1 \\
        Our Model with SIE-2 & 72.5 & 74.7 & 64.6 & 70.6 \\
        Our Model with SIE-9 & \textbf{72.8} & \textbf{75.3} & 66.0 & \textbf{71.9} \\
        Our Model with SIE-18 & 61.4 & 64.8 & 50.4 & 57.3 \\
        \hline
    \end{tabular}
    \caption{Comparison of Rank-1 and mAP scores for VIS-to-IR and IR-to-VIS modes on LLCM Dataset. *All numbers are reported without re-ranking to ensure a fair comparison. }
    \label{tab:reid_comparison}
\end{table*}

\textbf{Ablation Study:}

In \Cref{tab:ablation}, we present the results of an ablation study evaluating the impact of SIE, triplet loss, and GeM pooling. The results demonstrate that SIE consistently improves performance over the baseline in both VIS-to-IR and IR-to-VIS matching modes. Additionally, the study highlights that incorporating triplet loss alongside cross-entropy loss is crucial for achieving better discriminative feature learning. Finally, we analyze the effect of integrating GeM (Generalized Mean Pooling), showing that combining GeM-pooled outputs from patch embeddings with local and global features enhances the model's representational capacity specifically in the IR-to-VIS matching mode.

\textbf{Feature Space Visualizations:}

\begin{figure}[h!]
	\centering
	\includegraphics[width=0.9\linewidth]{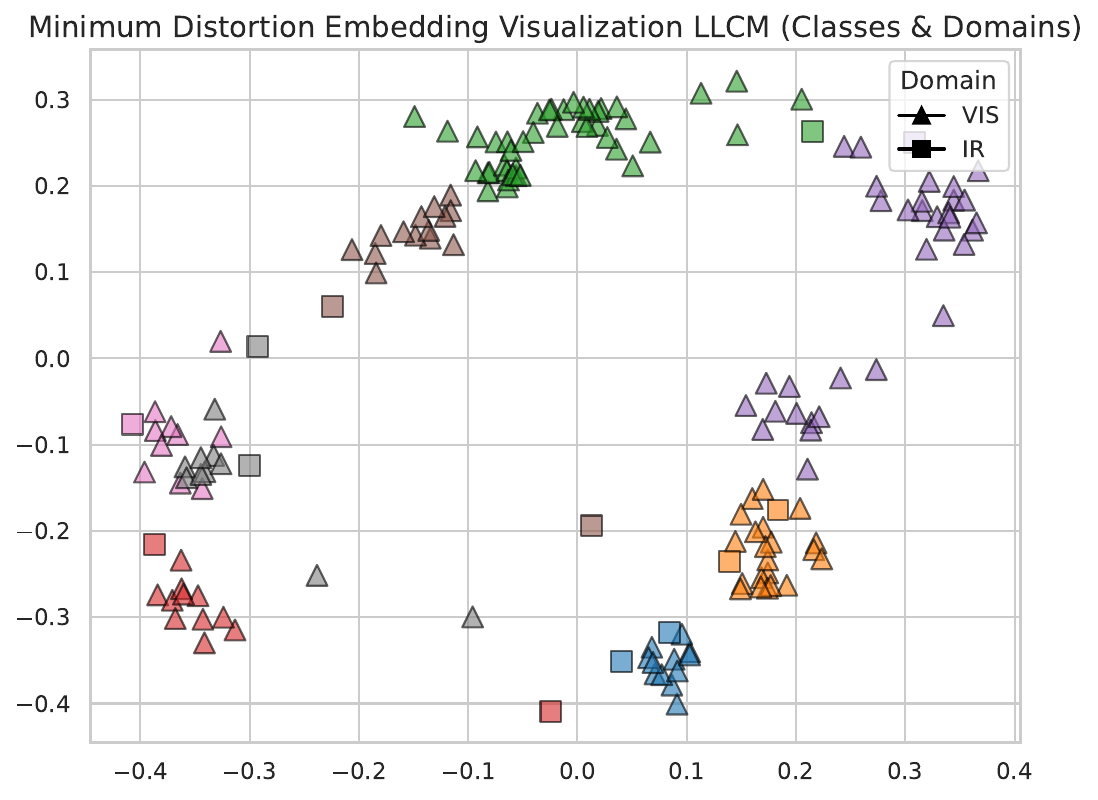}
	\caption{Minimum Distortion Embedding of LLCM features with the SIE-9 model}
	\label{fig:mde1}
\end{figure}

\begin{figure}[h!]
	\centering
	\includegraphics[width=0.85\linewidth]{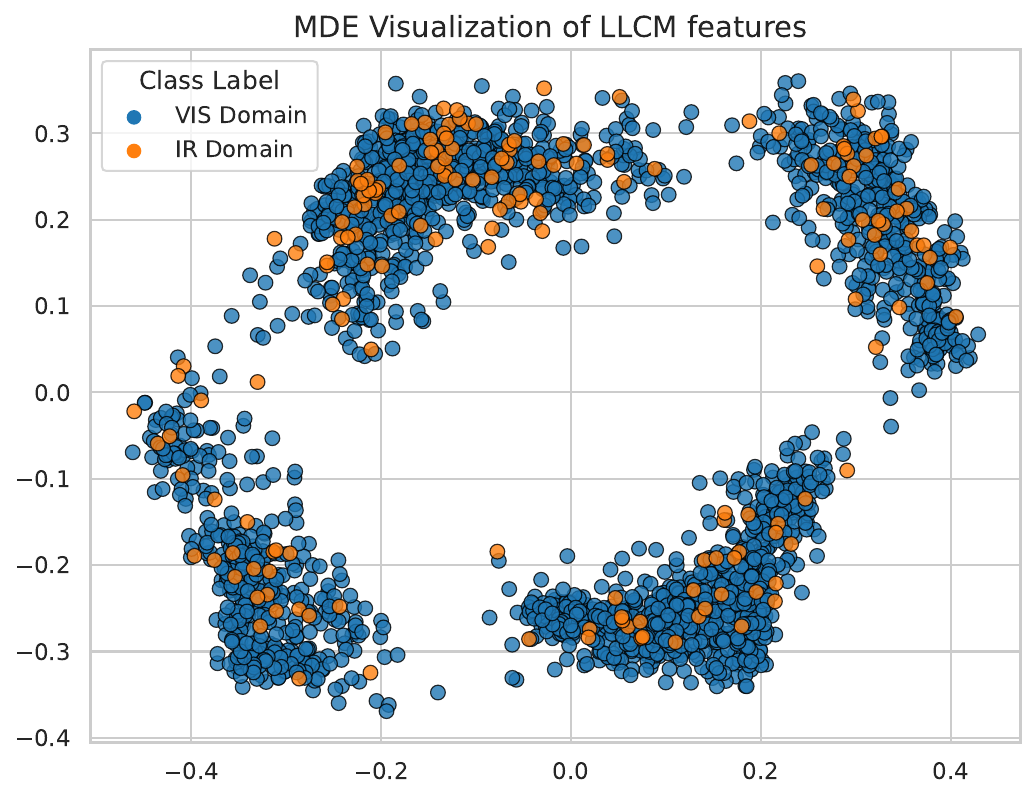}
	\caption{Minimum Distortion Embedding of LLCM features from 100 different subjects with the SIE-9 model}
	\label{fig:mde4}
\end{figure}

We now examine the feature space using Minimum Distortion Embedding (MDE) visualizations \cite{agrawal2021minimum}. \Cref{fig:mde1} illustrates the features extracted from test images of eight subjects, spanning both VIS and IR domains, using the SIE-9 model. Remarkably, the figure demonstrates that even without explicitly encoding domain information, the features from the IR and VIS domains are well-clustered together, while maintaining clear intra-class separation. This suggests that encoding camera information alone is effective in achieving robust cross-modal alignment. \Cref{fig:mde4} presents the MDE visualization of features extracted from 100 different subjects spanning both VIS and IR domains. The visualization demonstrates that the learned features exhibit strong robustness to domain shifts, with minimal separation between VIS and IR samples.

\begin{table*}[t!]
\renewcommand{\arraystretch}{1.5}
    \centering
    
    \begin{tabular}{l|c|c|c|cc|cc}
        \hline
        \multirow{2}{*}{\textbf{Model}} & \multirow{2}{*}{\textbf{SIE}} & \multirow{2}{*}{\textbf{Triplet Loss}} & \multirow{2}{*}{\textbf{GeM}} & \multicolumn{2}{c|}{\textbf{VIS to IR}} & \multicolumn{2}{c}{\textbf{IR to VIS}} \\
        
        & & & & Rank-1 & mAP & Rank-1 & mAP \\
        \hline
        Baseline (no SIE, Triplet, no GeM Pool) & \ding{55} & \ding{51} & \ding{55} & 70.5 & 73.2 & 63.9 & 70.1 \\
        + SIE-2 & \ding{51} & \ding{51} & \ding{55} & \textbf{72.5} & \textbf{74.7} & 64.6 & 70.6 \\
        + SIE-2 - Triplet Loss & \ding{51} & \ding{55} & \ding{55} & 60.3 & 63.3 & 55.5 & 61.3 \\
        + SIE-2 + GeM Pool & \ding{51} & \ding{51} & \ding{51} & 71.8 & 74.2 & \textbf{64.9} & \textbf{71.0} \\
        \hline
    \end{tabular}
    \caption{Ablation study on the impact of SIE, triplet loss, and GeM pooling.}
    \label{tab:ablation}
\end{table*}

\subsubsection{IJB-MDF}
\Cref{tab:multi-domain-vis500} presents the performance of models trained on short-range, long-range and mixed-range datasets, evaluated on the long-range protocol, while \cref{tab:multi-domain-vis300} presents the performance with the short-range evaluation protocol. For simplicity, we do not include results with the more challenging mixed-range test protocols. The first three rows represent models trained without SIE. Comparing the first three rows, the results indicate that on the long-range test protocol (\cref{tab:multi-domain-vis500}), the mixed-range model outperforms the long- and short-range models, while the short-range model performs best on the short-range protocol (except on the SWIR domain).

Incorporating domain information into SIE improves the performance of the mixed-range model across the SWIR and MWIR domains, on the long-range evaluation protocol (\cref{tab:multi-domain-vis500}). Interestingly, encoding both domain and range information (SIE-8) leads to a drop in performance, suggesting that encoding the range information using SIE is not the ideal strategy. 

These results highlight that pretrained body recognition transformer models struggle to adapt to short-range domains, even when the range information is encoded through side embeddings.

\begin{table*}[t!]
\renewcommand{\arraystretch}{1.5}
\centering
	\small
	\begin{tabular}{l|c |c |c}
		\hline
		\multirow{2}{*}{Network} & \multicolumn{3}{c}{Probe Domains (Gallery: VIS-500m)} \\
		\cline{2-4}
		& SWIR-30m & MWIR-30m & LWIR-30m \\
		\hline
		Long-Range Model & 69.7 & 67.8 & 64.6 \\
		Short-Range Model & 52.1 & 45.2 & 42.6 \\
		Mixed-Range Model & 75.1 & 68.2 & \textcolor{red}{67.2} \\
		\hline
		Mixed-Range Model SIE-4 & \textcolor{red}{76.3} & \textcolor{red}{71.8} & 65.4 \\
		Mixed-Range Model SIE-8 & 74.3 & 69.6 & 60.5 \\
		\hline
	\end{tabular}
	\caption{Performance (Rank-1 \%) comparison across different networks with gallery domain VIS-500m and various probe domains. Red values indicate best performances in respective columns.}
	\label{tab:multi-domain-vis500}
\end{table*}

\begin{table*}[t!]
\renewcommand{\arraystretch}{1.5}
\centering
	\small
	\begin{tabular}{l|c |c |c}
		\hline
		\multirow{2}{*}{Network} & \multicolumn{3}{c}{Probe Domains (Gallery: VIS-300m)} \\
		\cline{2-4}
		& SWIR-15m & MWIR-15m & LWIR-15m \\
		\hline
		Long-Range Model & 54.2 & 46.9 & 45.9 \\
		Short-Range Model & 69.7 & \textcolor{red}{67.2} & \textcolor{red}{63.9} \\
		Mixed-Range Model & 70.4 & 58.2 & 59.6 \\
		\hline
        
		Mixed-Range Model SIE-4 & \textcolor{red}{71.2} & 60.6 & 59.6 \\
		Mixed-Range Model SIE-8 & 65.5 & 56.9 & 56.6 \\
		\hline
	\end{tabular}
	\caption{Performance (Rank-1 \%) comparison across different networks with gallery domain VIS-300m and various probe domains. Red values indicate best performances in respective columns.}
	\label{tab:multi-domain-vis300}
\end{table*}

\textbf{MDE Visualizations:}

\begin{figure}[h!]
	\centering
	\includegraphics[width=0.85\linewidth]{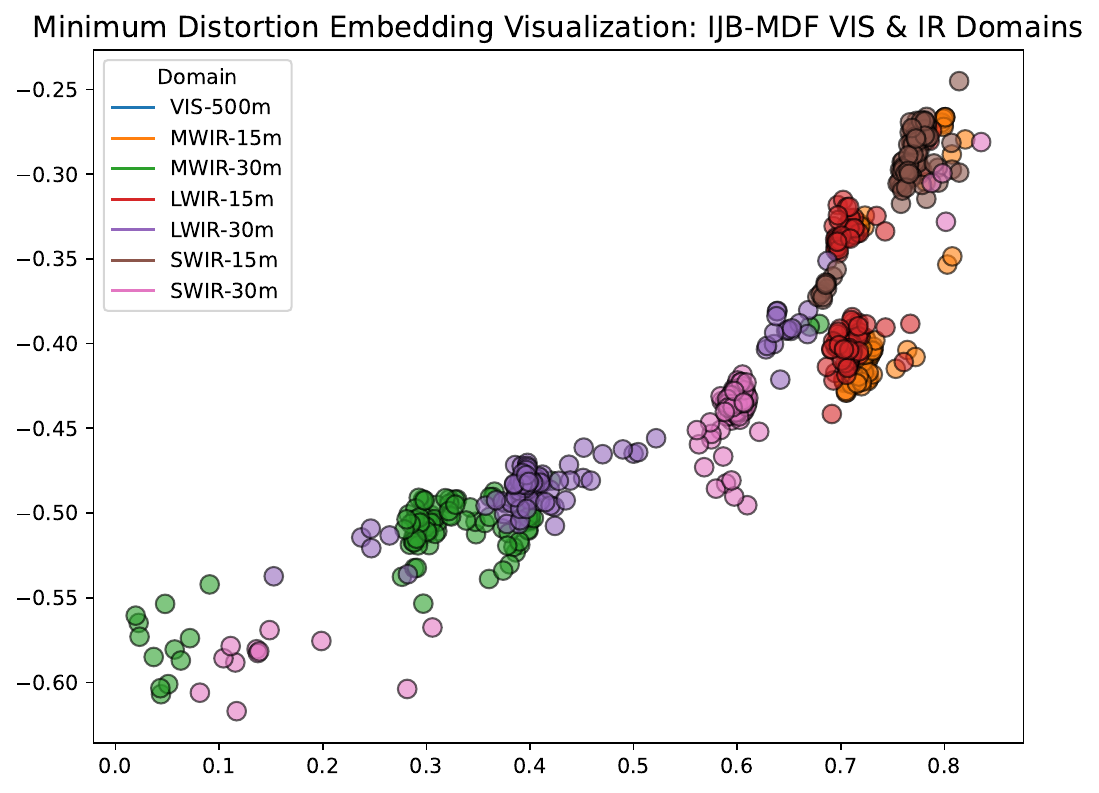}
	\caption{Minimum Distortion Embedding of MDF features of a single subject with the SIE-4 model}
	\label{fig:mde2}
\end{figure}

\Cref{fig:mde2} shows the MDE visualization of features extracted from test images of a single subject, spanning VIS-500m, as well as short-range and long-range IR domains, using the SIE-4 model. The visualization indicates that short-range and long-range features tend to cluster more closely together, regardless of the domain. This observation suggests that cross-range matching is more challenging than cross-spectral matching, for this model.

\begin{figure}[h!]
	\centering
	\includegraphics[width=0.85\linewidth]{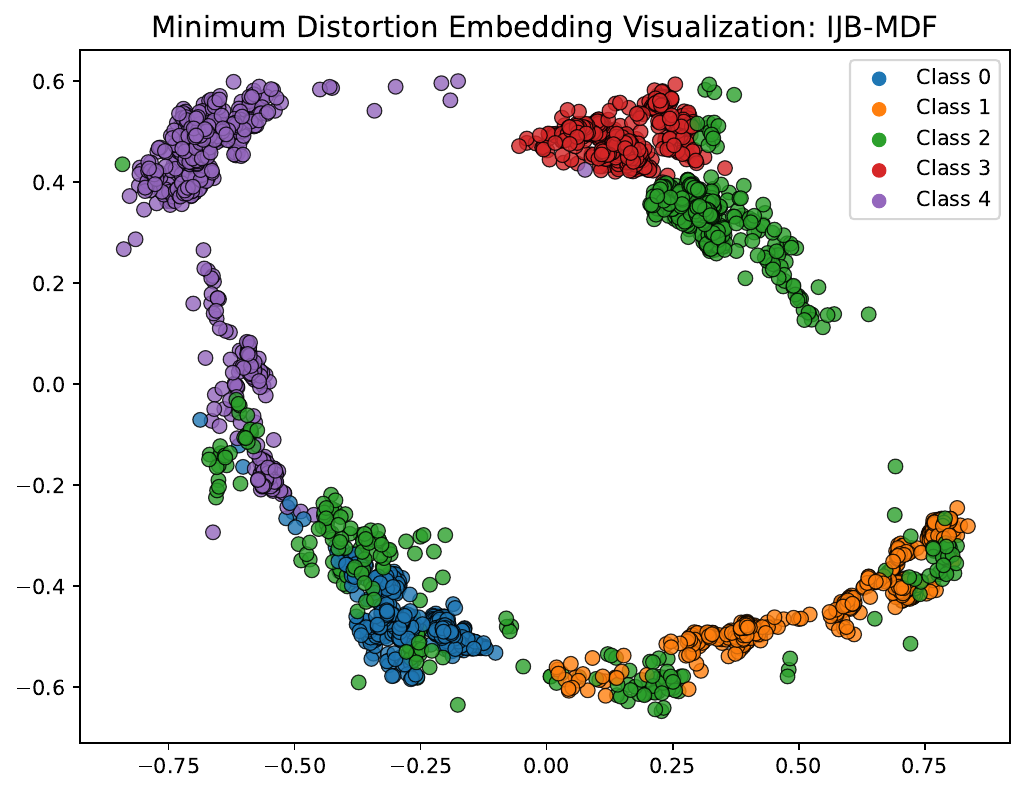}
	\caption{Minimum Distortion Embedding of MDF features of 5 different subjects with the SIE-4 model}
	\label{fig:mde3}
\end{figure}

\Cref{fig:mde3} displays the MDE visualization of features extracted from test images of five different subjects, covering both VIS and IR domains. For most subjects, the features form two distinct clusters, which, in light of insights from \cref{fig:mde2}, we hypothesize correspond to the short-range and long-range domains.

\section{Conclusion}
In this paper, we demonstrated that a ViT model pretrained solely on visible imagery can achieve state-of-the-art performance on the VI-ReID LLCM dataset by encoding camera information through Side Information Embeddings. Additionally, we investigated the impact of scale misalignment in cross-spectral body recognition, specifically due to variations in imaging range, using the IJB-MDF dataset. Our experiments demonstrate that even a state-of-the-art ViT model struggles to effectively match images captured at varying distances. This is primarily due to the fact that person recognition datasets typically consist of full-body images, causing body recognition models pretrained on these datasets to be inherently biased toward long-range imagery. As a result, these models fail to produce robust embeddings for short-range images, emphasizing the need for approaches that can facilitate transfer learning across scale variations in cross-spectral body recognition.

\section{Acknowledgements}
SH and RC are supported by the BRIAR project. This research is based upon work supported in part by the Office of the Director of National Intelligence (ODNI), Intelligence Advanced Research Projects Activity (IARPA), via [2022-21102100005]. The views and conclusions contained herein are those of the authors and should not be interpreted as necessarily representing the official policies, either expressed or implied, of ODNI, IARPA, or the U. S. Government. The US. Government is authorized to reproduce and distribute reprints for governmental purposes
notwithstanding any copyright annotation therein.
